\documentclass[10pt,twocolumn,letterpaper]{article}

\usepackage[pagenumbers]{wacv} 

\usepackage{array}
\usepackage[table]{xcolor}
\usepackage{graphicx}
\usepackage{amsmath}
\usepackage{amssymb}
\usepackage{booktabs}
\usepackage{float}
\usepackage{comment}
\usepackage[accsupp]{axessibility}
\usepackage{tikz}

\usepackage[pagebackref,breaklinks,colorlinks]{hyperref}

\usepackage[capitalize]{cleveref}
\crefname{section}{Sec.}{Secs.}
\Crefname{section}{Section}{Sections}
\Crefname{table}{Table}{Tables}
\crefname{table}{Tab.}{Tabs.}

\newcommand{\joh}[1]{~\textcolor{blue}{johannes:\ #1}}

\renewcommand{\joh}[1]{}

\def\wacvPaperID{503} 
\def\confName{WACV}
\def\confYear{2025}

\begin{document}

\title{Distillation of
Diffusion Features for
Semantic Correspondence}

\author{
Frank Fundel \and Johannes Schusterbauer \and Vincent Tao Hu \and Björn Ommer\\
CompVis@LMU Munich, MCML\\
\url{https://compvis.github.io/distilldift}\\
}
\maketitle


\begin{abstract}
   Semantic correspondence, the task of determining relationships between different parts of images, underpins various applications including 3D reconstruction, image-to-image translation, object tracking, and visual place recognition.
Recent studies have begun to explore representations learned in large generative image models for semantic correspondence, demonstrating promising results. Building on this progress, current state-of-the-art methods rely on combining multiple large models, resulting in high computational demands and reduced efficiency.
In this work, we address this challenge by proposing a more computationally efficient approach.
We propose a novel knowledge distillation technique to overcome the problem of reduced efficiency. We show how to use two large vision foundation models and distill the capabilities of these complementary models into one smaller model that maintains high accuracy at reduced computational cost.
Furthermore, we demonstrate that by incorporating 3D data, we are able to further improve performance, without the need for human-annotated correspondences.
Overall, our empirical results demonstrate that our distilled model with 3D data augmentation achieves performance superior to current state-of-the-art methods while significantly reducing computational load and enhancing practicality for real-world applications, such as semantic video correspondence. Our code and weights are publicly available on our project page.
\vspace{-1em}

\begin{figure}
    \centering
    \includegraphics[width=.49\textwidth]{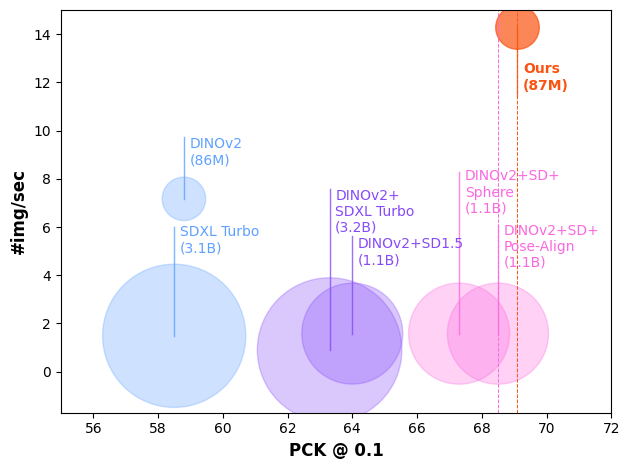}
    \caption{\textbf{Our method achieves better performance and throughput with less parameters on the SPair-71k dataset.} The circle size represents number of parameters. For more details, see \cref{tbl:results_performance}.}
    \label{fig:intro_results}
    \vspace{-1em}
\end{figure}
\end{abstract}
\section{Introduction}
Semantic correspondence involves finding relationships between different regions across two images, enabling a wide range of applications such as 3D reconstruction \cite{kokkinos2021point}, image-to-image translation \cite{tumanyan2022splicing}, object tracking \cite{gao2022aiatrack}, video segmentation \cite{wang2019learning}, pose estimation \cite{Shugurov_2022, huang2022}, explainable AI \cite{nguyen2023visual} and visual place recognition \cite{keetha2023anyloc}. Traditional approaches often rely on handcrafted features like SIFT \cite{SIFT} and SURF \cite{SURF}, which, despite their robustness, have limitations in capturing semantic meaning. With the advent of deep learning, Convolutional Neural Networks \cite{zeiler2013visualizing, han2017scnet, rocco2017convolutional, min2020learning} and Vision Transformers \cite{kim2018recurrent, Kim_2022_CVPR, cho2021cats, jiang2021cotr, cho2022cats} have revolutionized the field, providing powerful methods to extract semantically rich features from images in various ways.
Since finding semantic correspondences is challenging due to the need for extensive world knowledge, the transition to large-scale supervised deep learning was hindered by the limited availability of ground-truth annotations. To address this issue, recent efforts have focused on weakly-supervised \cite{wang2020learning} and self-supervised methods to extract per-pixel descriptors, most notably DINO \cite{amir2022deep, gupta2023asic}.
Diffusion Models (DMs) have shown great capabilities in generating high-quality images, due to their large-scale training paradigm which has also led them to learn rich world representations, useful for downstream tasks such as semantic correspondence \cite{hedlin2023unsupervised, tang2023emergent}.

Since a previous work ``A Tale of Two Features"~\cite{zhang2023tale} demonstrated that those two models are complementary in their features, current state-of-the-art approaches \cite{zhang2024telling, mariotti2023improving} all rely on combining two large vision foundation models. This comes with several limitations. First, diffusion models often need to be run multiple times to incorporate different timesteps, leading to increased computational complexity and longer inference times. Second, these combined models are typically very large networks with a substantial number of parameters, which require significant computational resources and memory. Third, managing these combined models introduces numerous additional hyperparameters, such as timestep(s), layer(s) and prompt, and mixing weights, complicating further training. This becomes particularly challenging when calculating correspondences in videos or in other real-time applications such as semantic video correspondence. Additionally, the increased number of parameters reduces computational efficiency, posing challenges for server-less applications.

Distillation~\cite{hinton2015distilling} is a well-known method broadly applied in various tasks like classification~\cite{li2021data}, segmentation~\cite{liu2019structured}, self-supervised learning~\cite{grill2020bootstrap} and generative models' sampling acceleration~\cite{salimans2022progressive}. 
This naturally raises a question: can we also apply distillation on top of the "Tale of Two Features"~\cite{zhang2023tale}, and ideally squeeze the parameter burden further to boost inference speed in the task of semantic correspondence?

We propose to leverage knowledge distillation for semantic correspondence to reduce the computational load of existing off-the-shelf methods, without retraining a full model. In summary, we make the following contributions:
\begin{itemize}
    \item We propose a parameter-efficient approach to distill the semantic correspondence capabilities of two complementary large vision foundation models into a more compact and efficient model.
    \item We additionally propose a novel fine-tuning protocol to further boost the model's performance by incorporating 3D data augmentation. This approach not only achieves new state-of-the-art results but also allows the model to improve without the need for labeled data.
    \item We validate the effectiveness of our method on three canonical benchmark datasets, demonstrating state-of-the-art performance on semantic correspondence with significantly reduced inference time.
\end{itemize}

\section{Related Work}
\textbf{Visual Correspondence.}
Visual correspondence refers to the identification of meaningful associations between different points or regions in images.
Traditional methods for geometric correspondence, such as SIFT \cite{SIFT, SIFT-Flow} and SURF \cite{SURF}, involve extracting hand-crafted visual descriptors and matching them across images. Later, it was discovered, that semantic correspondence also emerges in deep Convolutional Neural Network (CNN) features \cite{long2014convnets, zeiler2013visualizing, Zhou_2016_CVPR, gonzalezgarcia2017semantic} trained on other tasks such as classification. Subsequently, specifically designed architectures and algorithms were proposed to tackle the task of semantic correspondence using deep features of CNNs \cite{choy2016universal, kim2017fcss, 8451325, lee2019sfnet, rocco2018neighbourhood, rocco2020efficient, li2020correspondence, Lee_2021_CVPR, kim2018recurrent, Kim_2022_CVPR} and recently also transformers \cite{cho2021cats, jiang2021cotr, cho2022cats}.
However, finding semantic correspondences is a much more challenging task that requires vast knowledge about the world. Thus, the transition to large-scale supervised deep learning has been impeded by the limited availability of ground-truth correspondence annotations. As a solution to this issue, recent efforts have emerged in the form of weakly-supervised \cite{wang2020learning} and self-supervised \cite{NEURIPS2020_e2ef524f} methods to generate per-pixel (dense) descriptors, most notably DINO \cite{amir2022deep, gupta2023asic}.
DINO is a large Vision~Transformer~(ViT)~\cite{vit} trained using a self-distillation approach, where the model learns to predict its own output under different augmentations, effectively leveraging large amounts of unlabeled data to generate high-quality dense descriptors.
Generative models need to possess extensive knowledge about the world to synthesize new data accurately.
Therefore, the internal representations of large generative models, such as Generative Generative Adversarial Networks~(GANs)~\cite{gan}, can be effectively utilized for identifying visual correspondences within specific image categories \cite{mu2022coordgan}. Similarly, internal representations of Diffusion Models (DMs) recently demonstrated their potential in various downstream tasks.\\

\textbf{Diffusion Model Representations.}
\label{sec:diffrep}
Diffusion Models have significantly advanced the field of image generation, achieving state-of-the-art performance \cite{diffusionbeatgans, latentdiffusion,schusterbauer2024boosting} and their applications extend beyond image creation to various vision tasks such as image segmentation \cite{amit2022segdiff, baranchuk2022labelefficient, chen2023generalist, jiang2018difnet, tan2023semantic, wolleb2021diffusion, tan2023diffss}, object detection \cite{chen2023diffusiondet}, image classification \cite{li2023diffusion, mukhopadhyay2023diffusion}, image editing \cite{hulfm,kawar2023imagic, choi2023customedit, shi2023dragdiffusion, couairon2022diffedit} and monocular depth estimation \cite{duan2023diffusiondepth, saxena2023monocular, shao2023monodiffusion, gui2024depthfm}.
Recently, there has been a growing interest in exploring the underlying representations of these models, particularly in how they can be leveraged for various downstream tasks~\cite{fuest2024diffusion}. Specifically, several works have demonstrated that the intermediate cross-attention maps of a pre-trained diffusion model can be used for text-based image editing \cite{tumanyan2022plugandplay, hertz2022prompttoprompt}, zero-shot segmentation \cite{xu2023openvocabulary, baranchuk2022labelefficient, wu2023diffumask} and depth estimation \cite{zhao2023unleashing}.
These methods work by aggregating intermediate feature maps, self- or cross-attention maps from a hand-selected subset of layers and timesteps during the forward diffusion process.\\

\textbf{Diffusion Models for Semantic Correspondence.}
\label{sec:diffusion4semanticcorr}
While the applicability of diffusion model representations in various computer vision tasks has been well-established, their use in the domain of semantic correspondence is still an emerging area of research with only a few recent works.
Hedlin et al. and Li et al. \cite{hedlin2023unsupervised, li2024sd4match} focus on optimizing the Diffusion Model's prompt embeddings to find correspondences across images. However, due to the test-time optimization of the prompt embedding and iterative sampling of the diffusion model, the overall computational efficiency hinders it from being applied to further tasks, such as fine-tuning.
Tang et al. \cite{tang2023emergent} present a straightforward approach that utilizes cosine similarity of the intermediate features of two images at a specific timestep and layer to find correspondences using nearest neighbor lookup.
Luo et al. \cite{luo2023diffusion} introduce a technique that leverages a learned feature aggregation network for combining features over multiple layers and timesteps.
``A Tale of Two Features" \cite{zhang2023tale} demonstrate that the features of a Diffusion Model and those of DINOv2 complement each other and that their combined performance is greater than their individual.
Additionally, recent studies focus on overcoming the issue of left-right ambiguity \cite{mariotti2023improving, zhang2024telling}.
However, while all these methods provide accurate semantic representations, they still rely on aggregating the extracted features of two large vision foundation models, resulting in high computational demands and reduced efficiency.\\

\textbf{Knowledge Distillation.}
\label{sec:distillation}
Knowledge Distillation is a process where knowledge is transferred from one model, that is typically larger and more complex, to another model, that is usually smaller and more efficient \cite{hinton2015distilling}. 
It has been successfully applied to (large) language models \cite{sanh2020distilbert, gu2023knowledge}, speech recognition models \cite{gandhi2023distilwhisper} and even diffusion models \cite{salimans2022progressive, meng2023distillation, song2023consistency, gu2023boot, shao2023monodiffusion} in various ways.
There are three kinds of knowledge: response based \cite{hinton2015distilling, Meng_2019} (the output layer's logits), feature based \cite{bengio2014representation, romero2015fitnets, zagoruyko2017paying} (the intermediate feature maps) and relation based \cite{Yim_2017_CVPR} (the relationships between different layers or data samples).
Different techniques can be utilized to perform distillation e.g. multi-teacher distillation \cite{sau2016deep, park2019feed}, where multiple teacher models are combined to train a single student model. Pseudo-labeling for distillation \cite{kim-rush-2016-sequence} involves the generation of pseudo-labels for previously unlabeled data by the teacher model.
Some works use cross-entropy between the teacher and student outputs \cite{hinton2015distilling, sanh2020distilbert}. Others aim to minimize the reverse KLD, which measures how two probability distributions differ \cite{gu2023knowledge, gandhi2023distilwhisper}. For Diffusion Models, knowledge distillation is primarily applied to reduce the number of diffusion steps needed to generate high-fidelity images, with multiple works utilizing a weighted MSE as the objective \cite{salimans2022progressive, meng2023distillation, song2023consistency, gu2023boot}.
One work \cite{Li_2021_CVPR} specifically focuses on leveraging distillation for semantic correspondence. However, they aim to distill knowledge from a probabilistic model which mimics a group of models into a single static one, whereas our goal is to efficiently distill the knowledge of multiple foundation model teachers into a single smaller model. Furthermore, they utilize synthetic image pairs by augmenting a single image, whereas we make use of real image pairs from a multi-view image dataset.
We utilize multi-teacher distillation to distill the semantic correspondence capabilities of two large models into a single model, by aligning their predicted similarity distributions.

\section{Method}
\textbf{Problem Formulation.}
Semantic correspondence can be defined as follows. For a given pair of images $I_1, I_2$ and a query keypoint $p_1 \in I_1$, we want to find the corresponding target point $p_2 \in I_2$ with the highest semantic similarity i.e. points of different objects that share semantic meaning. By extracting feature maps $\mathcal{F}_1, \mathcal{F}_2$ from both images, a corresponding target point for $p_2$ can be obtained with:
\begin{equation}
    p_2 = \arg\min_p d(\mathcal{F}_1(p_1), \mathcal{F}_2(p)),
\end{equation}
where the distance metric $d(\cdot, \cdot)$ is defined as cosine similarity $\mathrm{sim}(\cdot,\cdot)$. 
Current state-of-the-art methods~\cite{zhang2023tale, zhang2024telling, mariotti2023improving} rely on combining two large vision foundation models, specifically a Vision Transformer and a Diffusion Model. This introduces numerous hyperparameters (e.g., timesteps, layers, prompts, mixing weights), increasing the complexity, and number of parameters. As a result, fine-tuning is difficult, VRAM requirements are higher, and computational efficiency is reduced, challenging server-less or real-time deployment.
In this work, we address these issues by distilling information from these two large vision foundation models into a single model. This approach significantly reduces complexity and decreases the number of parameters, thereby improving efficiency without compromising performance. Furthermore, we show that by incorporating 3D data from a multi-view image dataset, we can fine-tune our model without the expense of human-annotated data.

\begin{figure*}
    \centering
    \includegraphics[width=1.0\textwidth]{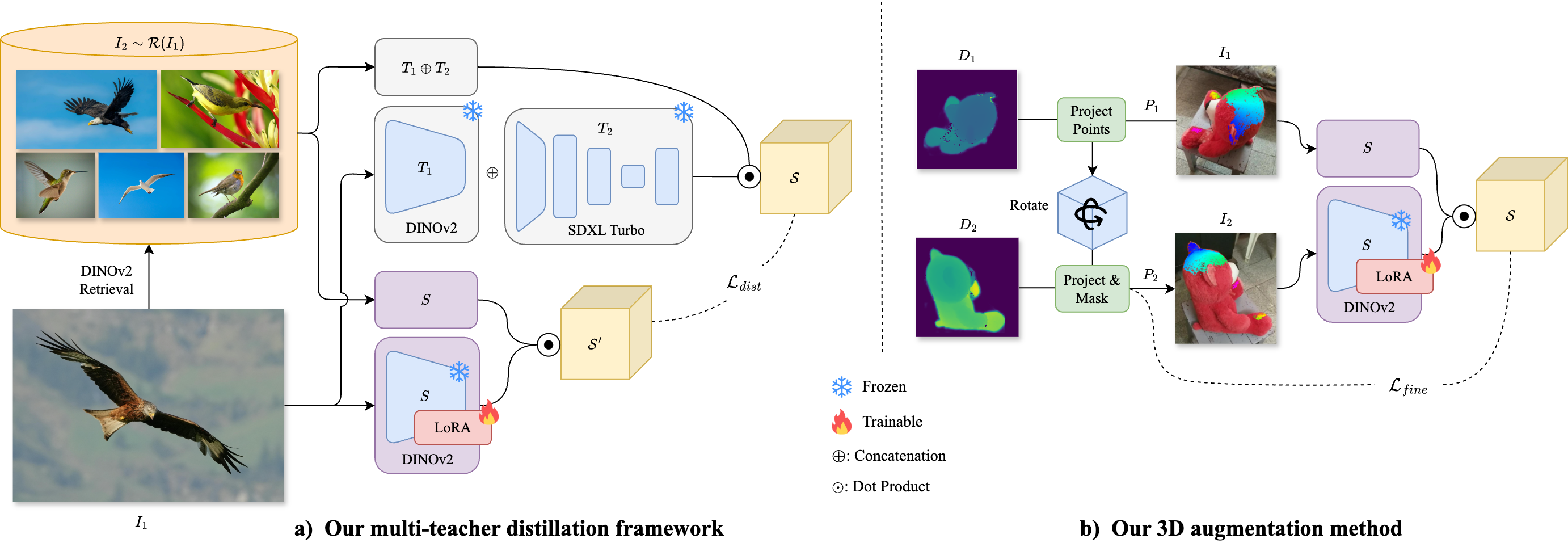}
    \caption{\textbf{Illustration of our multi-teacher distillation framework (a) and 3D data augmentation method (b).} We distill two complementary models, DINOv2 and SDXL Turbo, into one single and more efficient model. Using unsupervised 3D data augmentation we further refine our distilled model to achieve new state-of-the-art in both throughput and performance.}
    \label{fig:method}
    \vspace{-1em}
\end{figure*}

\subsection{Preliminaries}
We leverage two major models that have shown strong performance in semantic correspondence:
Stable Diffusion \cite{latentdiffusion} is a text-conditioned Latent Diffusion Model (LDM) known for its ability to synthesize realistic images, leading to highly informative internal feature representations \cite{tang2023emergent}.
DINOv2 \cite{oquab2023dinov2} is a self-supervised Vision Transformer that excels in capturing useful features through self-supervised contrastive learning.
Combining these two models offers complementary strengths, which are rooted in their distinct learning paradigms \cite{zhang2023tale}. While LDMs learn to generate images with a holistic understanding of the world, capturing both global and local information, DINO focuses on extracting meaningful features with invariance to spatial information, i.e. local position and global orientation, due to its input augmentations during training.

\subsection{Framework}
\textbf{Multi-Teacher Distillation.}
\label{sec:method_distillation}
Our method aims to distill information from two large vision foundation models into a single student model that provides fast inference with accurate predictions. Therefore, we employ multi-teacher distillation with two teacher models $T_1, T_2$ and one student model $S$, where $T_1$ is a ViT and $T_2$ is a diffusion model. To achieve our goal of parameter-efficient training, we initialize the student model with a pre-trained vision model and incorporate Low-Rank Adaptation (LoRA) \cite{lora}.
LoRA uses two smaller matrices, $A \in \mathbb{R}^{r \times k}$ and $B \in \mathbb{R}^{d \times r}$, whose dot product $\Delta W = BA$ matches the size of the model’s initial weight matrix $W \in \mathbb{R}^{d \times k}$, with its size controlled by $r \ll \min(d, k)$.
Following \cite{zhu2023melo, cui2024surgicaldino, bafghi2024parameter}, we apply LoRA only to the projection layers of the queries and values and keep all other parameters fixed:
\begin{equation}
\begin{aligned}
    Q &= W_Qx + B_Q A_Qx,\\
    V & = W_Vx + B_V A_Vx,\\
    K & = W_Kx,
\end{aligned}
\end{equation}
where  $W_Q, W_K$ and $W_V$ are the frozen projection layers and $B_Q, A_Q$ and $B_V, A_V$ are the trainable LoRA layers. The weights of $B_Q$ and $B_V$ are initialized with zeros. Besides parameter efficiency, this also mitigates the problem of catastrophic forgetting \cite{lora_forgets_less}.
We utilize the predicted feature maps of an image pair to obtain the similarity maps of the teacher and student. The objective of the distillation is to align these similarity distributions. Thus, our approach is fully unsupervised by leveraging pseudo-labeling instead of human-annotated correspondences.
Specifically, we pass an image $I_1$ to $T_1$ while we pass multiple noised versions $I_1 + \epsilon_t$ to $T_2$. We average the extracted feature maps of $T_2$ and concatenate them with the extracted feature map of $T_1$ to form a combined feature map $\mathcal{F}_1$. We input the same image $I_1$ to the student $S$ and extract the feature map $\mathcal{F}_1'$ from it.\\

However, manually selected image pairs are scarce. To exclude any supervision while still having image pairs that are beneficial for the distillation, we propose to use image retrieval during training. The goal is to find a set of semantically similar images $\mathcal{R}(I_1)$ for a given image $I_1$. After finding such a set, we sample and process a second image $I_2 \sim \mathcal{R}(I_1)$ in the same way as $I_1$, to obtain feature maps $\mathcal{F}_2$ and $\mathcal{F}_2'$.\\

We compute the cosine similarity of \emph{all vectors} of the teacher feature maps $\mathcal{F}_1, \mathcal{F}_2$ and the student feature maps $\mathcal{F}_1', \mathcal{F}_2'$ to obtain the similarity maps $\mathcal{S}$ and $\mathcal{S}'$ with shape $(H \times W) \times (H' \times W')$ using:
\begin{equation}
    \mathrm{sim}\left(\mathcal{F}_1, \mathcal{F}_2\right) = \mathcal{F}_1 \cdot \mathcal{F}_2^T = \mathcal{S}
\end{equation}

We apply softmax with temperature $\sigma_\tau(\cdot)$ to ensure that each descriptor has a sharp but distributed similarity over all other descriptors. We use cross-entropy as a dense objective over all $(W \times H)^2$ correspondences simultaneously. Hence, we define the final training objective as:
\begin{equation}
\begin{aligned}
    \mathcal{L}_{dist} &= \mathrm{CE}\left(\sigma_\tau(\mathcal{F}_1 \cdot \mathcal{F}_1'^T), \right.\left. \sigma_\tau(\mathcal{F}_2 \cdot \mathcal{F}_2'^T) \right) \\
    &= \mathrm{CE}\left(\mathcal{S}, \mathcal{S}'\right)
\end{aligned}
\label{eq:loss_dist}   
\end{equation}
Here, cross-entropy loss $\mathrm{CE}$ is defined as $\mathrm{CE}(P, T)=-\mathbb{E}_{P}[\log T]$
where $H$ is the cross-entropy of the distribution $T$ relative to a distribution $P$.
As a result, the student learns the approximate goal of dense image similarity instead of directly imitating the teacher's features and lowering the feature dimension.\\

\textbf{3D Data Augmentation.}
\label{sec:enhancement}
Because large and diverse annotated datasets for semantic correspondence are scarce, we utilize a large multi-view image dataset with corresponding depth maps, namely CO3D \cite{reizenstein21co3d}, to further enhance our distilled model in an unsupervised manner. Specifically, with two images of different viewing angles $I_1, I_2$, their corresponding depth maps $D_1, D_2$, and camera parameters $K_1, K_2$, we project all points of $I_1$ from screen view $s$ into world view $w$ using $D_1, K_1$ and project them back into screen view using $K_2$:
\begin{equation}
\begin{gathered}
    T = T_{s \rightarrow w}(K_1) \cdot T_{w \rightarrow s}(K_2),\\
    \hat{p} = \left(\begin{array}{c}p_x \\ p_y \\ D_1(p_x, p_y)\end{array}\right) \cdot T,
\end{gathered}
\end{equation}
where $T_{i \rightarrow j}(\cdot)$ is the transformation from coordinate system $i$ to $j$. Using the difference in $z$-axis of the projected depth $\hat{D}_2$ to $D_2$, we exclude points that are not visible after the transformation:
\begin{equation}
M(x, y) = \begin{cases} 
1 & \text{if } |D_2(x, y) - \hat{D}_2(x, y)| < \epsilon, \\
0 & \text{otherwise}
\end{cases}
\label{eq:3d_visible}
\end{equation}
where $M$ is the mask of mutually visible pixels and $\epsilon$ is the threshold parameter.
Thus, we obtain \emph{locally-dense} correspondences to fine-tune our model without any human annotations. Following \cite{li2023simsc}, we apply a $k \times k$ Gaussian kernel $g_k(\cdot)$ to the correspondence points, resulting in a $(N \times H \times W)$ sized correspondence map. Furthermore, we apply softmax with temperature to the predicted similarity map $\mathcal{S}$. This yields two distributions, where we utilize cross-entropy as the learning objective, which leads to the final fine-tuning loss:
\begin{equation}
    \mathcal{L}_{fine} = \mathrm{CE}\left(\sigma_{\tau}\left(\mathcal{F}_1 \cdot \mathcal{F}_2^T\right), \; g_k\left(G\right)\right)
\label{eq:loss_fine}
\end{equation}
\section{Experiments}

\begin{table*}
\centering
\begin{tabular}{llccccc}
\toprule
\textbf{Model} & \textbf{Input size} & \textbf{\#Params} & \textbf{\#img/sec} & \textbf{SPair-71k} & \textbf{PF-WILLOW} & \textbf{CUB-200} \\
\midrule
\textbf{Unsupervised} & & & & \\
\midrule
DINOv2 & $840 \times 840$ & $\mathbf{87M}$ & \underline{7.2} & 59.5 & 67.4 & 85.9 \\
\hspace{1em}
\raisebox{0.2em}{
\begin{tikzpicture}[>=stealth]
    \draw[->] (0,0) -- (0,-0.2) -- (0.2,-0.2);
\end{tikzpicture}}
smaller resolution & $434 \times 434$ & $\mathbf{87M}$ & \textbf{28.6} & 59.2 & 65.8 & 85.3 \\
DIFT \cite{tang2023emergent} & $768 \times 768$ & $1B$ & 1.5 & 57.7 & \underline{76.4} & 74.3 \\
DINOv2+SD1.5 \cite{zhang2023tale} & $960 \times 960$ & $1.1B$ & 0.4 & \underline{64.0} & 74.8 & \underline{86.3} \\
\textit{Ours} & $434 \times 434$ & $\mathbf{87M}$ & \textbf{28.6} & \textbf{65.1} & \textbf{77.4} & \textbf{89.4} \\
\midrule
\textbf{Weakly-supervised} & & & & \\
\midrule
DINOv2+SD+Sphere\cite{mariotti2023improving} & $960 \times 960$ & \underline{$1B$} & 0.5 & 67.3 & - & - \\
DINOv2+SD+Pose-Align\textsuperscript{\textdagger} \cite{zhang2024telling} & $960 \times 960$ & $\underline{1B}$ & 0.8 & \underline{69.6} & - & - \\
\textit{Ours}+Pose-Align\textsuperscript{\textdagger} & $434 \times 434$ & $\mathbf{87M}$ & \textbf{14.3} & \textbf{70.6} & - & - \\
\bottomrule
\end{tabular}
\caption{\textbf{Comparison of total parameter count, throughput (images per second) on a single \emph{NVIDIA A100 80GB} and performance on different datasets.} Performances are measured in $\text{PCK}_{\text{bbox}}$@0.1 \textit{per point}, averaged over all points for PF-WILLOW and CUB200, and averaged over classes for SPair-71k. \textsuperscript{\textdagger}: Pose-align can only be applied to datasets that include keypoint-specific labels. The best performances are \textbf{bold}, while the second-best are \underline{underlined}.}
\label{tbl:results_performance}
\vspace{-1em}
\end{table*}

\textbf{Datasets.} To validate the performance of our approach, we follow previous works \cite{zhang2023tale, zhang2024telling, hedlin2023unsupervised, luo2023diffusion, tang2023emergent, li2024sd4match} and evaluate on the following benchmark datasets.
SPair-71k \cite{min2019spair71k} consists of 70{,}958 annotated image pairs, each carefully selected to include a wide variety of changes in viewpoint, scale, occlusion level, and truncation.
PF-WILLOW \cite{ham2016proposal} consists of 100 different images and 900 image pairs from 5 categories.
CUB-200-2011 \cite{WahCUB_200_2011} contains 11{,}788 bird images of 200 categories, with 15 distinct part locations each.
COCO \cite{lin2015microsoft} is a large-scale object detection, segmentation, key-point detection, and captioning dataset and consists of 328,000 images from 80 categories.
CO3D \cite{reizenstein21co3d} is a large multi-view image dataset that includes camera parameters, depth maps and segmentation masks and consists of almost 19{,}000 videos from 50 COCO categories.\\

\textbf{Evaluation.} We use the Percentage of Correct Keypoints (PCK) to evaluate the performance of correspondence. In this metric, a predicted keypoint is considered to be correct if it lies within a radius of $\alpha \cdot max(h, w)$. We measure two variants of PCK: $\text{PCK}_{\alpha=0.1}\text{img}$ and $\text{PCK}_{\alpha=0.1}\text{bbox}$, where ``img" indicates that $h, w$ are the dimensions of the image, whereas ``bbox" indicates that $h, w$ are the dimensions of the object's bounding box, included in the corresponding dataset.\\

Similar to \cite{zhang2024telling}, we apply soft-argmax on a pre-defined window around the predicted target point (window soft-argmax \cite{bulat2021subpixel}) during inference.
Following \cite{zhang2024telling}, we horizontally flip the source image and calculate the global similarity between the target image and both the original and flipped source images. We choose the image with higher similarity for predicting point correspondences with the target image (pose-align) to mitigate the problem of left-right ambiguity. This method uses point labels e.g. ``left paw", ``right paw", only available in SPair-71k to switch repetitive points after the flipping process, therefore we consider this method to be weakly supervised.\\

\textbf{Implementation Details.} We employ two teacher models $T_1, T_2$, where $T_1$ is a Vision Transformer and $T_2$ is a Diffusion Model. We use specific variants of DINOv2 and Stable Diffusion, namely DINOv2 B/14 with registers for the first teacher $T_1$ and the student $S$, and SDXL Turbo for the second teacher $T_2$.
Using DINOv2 as a prior for the student greatly enhances
the efficiency of our proposed method and reduces computational demands.
We extract the features of $S$ and $T_1$ from layer $11$, while we extract the features of $T_2$ from layer $1$. For the ensemble method, we average the feature maps of the diffusion teacher $T_2$ of multiple timesteps $t \in {51, 101, 151, 201}$. We set the rank for the LoRA of $S$ to $r = 8$ and the softmax temperature to $\tau = 0.01$.
To align the shapes of all feature maps, we set the size of the input images for $T_1$ and $S$ to $434 \times 434$, while for $T_2$ we set it to $980 \times 980$. Both $T_1$ and $T_2$ output feature maps with a spatial resolution of $31 \times 31$. This results in a combined feature map shape of $31 \times 31 \times (768 + 1280)$. The student predicts feature maps with shape $31 \times 31 \times 768$. 
For distillation, we use a dropout of $0.05$ on the LoRA layers, AdamW optimizer with a weight decay of $0.05$, an initial learning rate of $0.0001$, a step learning rate scheduler and train for $40$ epochs on a small subset (12.000 samples) of the COCO dataset. We use the top $k=10$ closest samples in the image retrieval process.
During the 3D data augmentation process, we use a Gaussian blur with kernel size $k = 7$ and a masking threshold of $\epsilon=0.01$.
During supervised fine-tuning however, a final linear head (2 layers) is added to the model and trained for $8$ epochs, while iteratively increasing the dropout to $0.1$.
We use the normalized feature matrices $\mathcal{F}_1, \mathcal{F}_2$ to eliminate the division by their lengths (dot-product similarity).\\

\begin{figure*}
    \centering
    \includegraphics[width=0.95\textwidth]{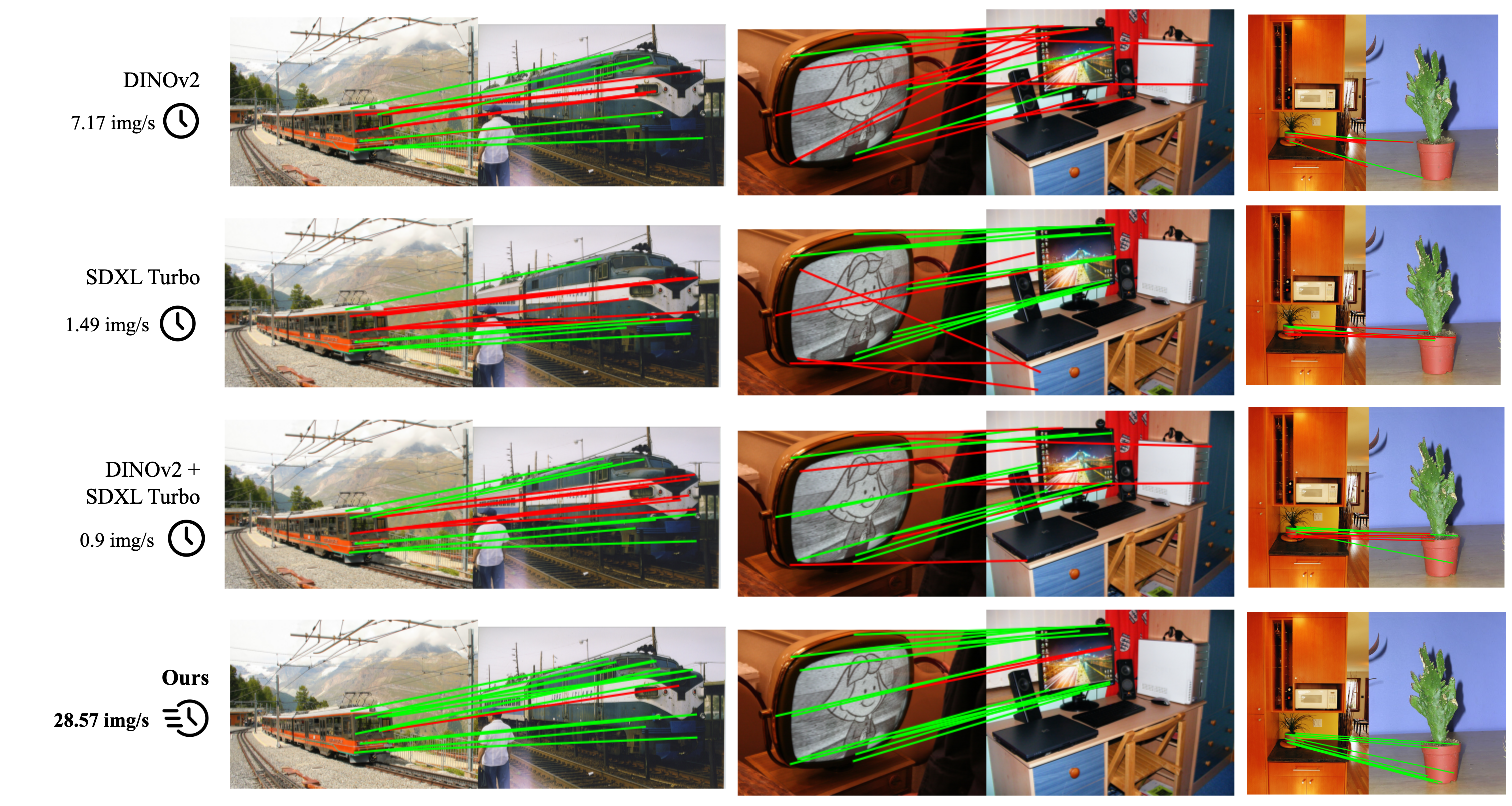}
    \vspace{-2mm}
    \caption{\textbf{Examples image pairs from SPair-71k with predicted correspondences of different methods.} \textcolor{green}{Green} indicates correct, while \textcolor{red}{red} indicates incorrect according to $\text{PCK}_{\text{bbox}}$@0.1. $(840 \times 840)$ was used as input resolution for DINOv2.}
    \label{fig:image_examples}
\end{figure*}

\subsection{Comparison with State-of-the-Art}
\label{sec:results}
\begin{figure*}
    \centering
    \includegraphics[width=0.9\textwidth]{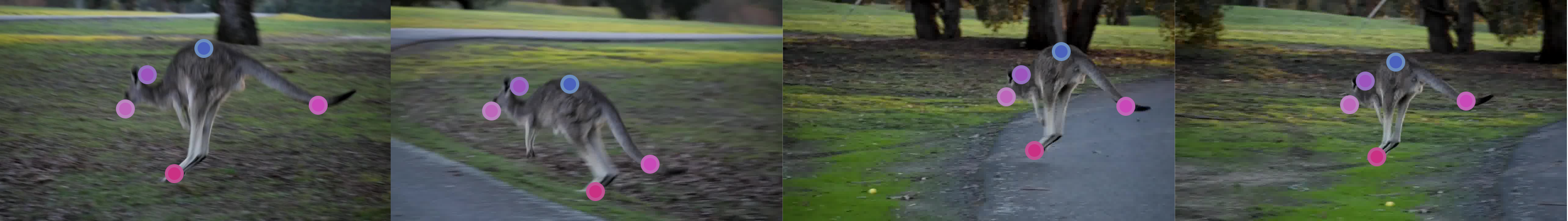}
    \vspace{-2mm}
    \caption{\textbf{Video semantic correspondence sample, showing accurate correspondences at a high frame rate.} We use source points on the first frame to calculate the corresponding points on all other frames at almost 30 FPS on an \emph{NVIDIA A100 80GB}.}
    \label{fig:video_example}
\end{figure*}

\cref{tbl:results_performance} and \cref{fig:image_examples} show that our distilled model performs superior in both weakly-supervised and unsupervised correspondence compared to the current state-of-the-art. Notably, it achieves these results with far fewer parameters and thus far faster throughput compared to previous models, while drastically reducing computational load. This demonstrates that we can successfully distill information from two large vision foundation models into a smaller student model, without sacrificing performance. The model's capability to work with smaller image sizes translates into fewer tokens processed, which additionally speeds up its performance. We were able to increase the number of images processed per second by a factor of $\textit{18}$, reduce the number of parameters by a factor of $\frac{1}{12}$ and improve $PCK_{bbox}$@0.1 by $\textit{0.6p}$ compared to the current state-of-the-art in weakly-supervised semantic correspondence. The naive approach of concatenating the feature maps of both models, as how it was introduced in "A Tale of Two Features" \cite{zhang2023tale, zhang2024telling, mariotti2023improving}, on the other hand, introduces significant overhead in terms of VRAM and throughput, making the switch to our model an attractive option for real-time applications like semantic video correspondence, as seen in \cref{fig:video_example}.

\subsection{Ablations}
\begin{figure}
    \centering
    \includegraphics[width=.4\textwidth]{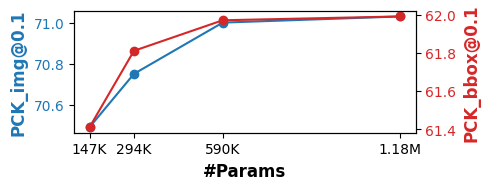}
    \vspace{-2mm}
    \caption{\textbf{Ablation of the rank parameter of LoRA, evaluated on SPair-71k.} Trained for 20 epochs on COCO with retrieval sampling. The \#Params correspond to the ranks: 4, 8, 16 and 32, respectively.}
    \label{fig:experiments_lora_rank}
\end{figure}

\textbf{Fine-Tuning Strategy.} A common parameter-efficient fine-tuning strategy is to add a linear head at the end of the model. However, a linear head, applied to each token separately, may only capture local structures. Thus, we ablate two different fine-tuning strategies for distillation: Linear head ($2$ layers) and LoRA \cite{lora} (rank $8$ on all layers). We exclude full fine-tuning, mainly to avoid catastrophic forgetting and the excessive training demands associated with it. As shown in \cref{tbl:experiments_peft}, the linear layers are not sufficient for this distillation and our method with LoRA performs superior. Additionally, we ablate the rank parameter of LoRA, and our findings suggest that a low rank is sufficient for distillation while providing a good trade-off between performance and number of trainable parameters, see \cref{fig:experiments_lora_rank}.
In \cref{tbl:experiments_enhance}, we demonstrate that our 3D data augmentation method further improves the model, proving that semantic correspondence can be enhanced without any cost-intensive human keypoint annotations.\\

\textbf{Point Sampling.} Instead of aligning all point similarities within a feature map, only a subset of influential points could be used for distillation. We evaluate three different point sampling techniques: (1) sample pixels using source point annotations, (2) sample pixels that are mutual nearest neighbors, and (3) sample from all pixels within the image. \cref{tbl:experiments_point_sampling} demonstrates, that the simplest approach of sampling all pixels leads to superior performance.\\

\textbf{Image Sampling.} Diverse pairs of images are crucial for effective distillation. However, SPair-71k only contains around 50 different images per category. Hence, we want to gradually shift to leveraging a large-scale dataset, instead of using manually selected image pairs provided by SPair-71k. We compare three alternative methods: (1) randomly combined images from the same category, (2) completely random combinations of images and (3) combining semantically similar images using retrieval. For the latter, we leverage DINOv2 embeddings to retrieve multiple similar target images for each source image. \cref{tbl:experiments_image_sampling} shows that our proposed technique (retrieval pairs) achieves comparable results even without any supervision on an independent dataset (COCO).\\

\textbf{Orthogonality to Previous Methods.} In \cref{tbl:experiments_enhance} we demonstrate, that current optimization techniques, such as an ensemble of varying diffusion timesteps $t \in {51, 101, 151, 201}$ instead of a single timestep \cite{tang2023emergent} during distillation, window soft-argmax, and pose-align \cite{zhang2024telling}, are also applicable to our model, further improving performance. Additionally, we apply our method to an independent dataset (COCO) to ensure minimal supervision and show that even in a completely unsupervised setting, our model is orthogonal to previous methods.
Finally, we fine-tune our model to show its capabilities in a supervised setting and also compare it to the state-of-the-art in \cref{tbl:experiments_finetune}, demonstrating comparable performance at much higher throughput, highlighting the effectiveness of our method.\\

\begin{figure}
    \setlength\tabcolsep{0.01pt}
    \newcommand{\imgwidth}{0.1\textwidth}
    \newcommand{\imagepng}[1]{
    \includegraphics[width=\imgwidth]{images/fg-bg/fg-bg_#1.png}
    }
    \centering
    \tiny
    \begin{tabular}{ccc}
        Image & DistillDIFT & DINOv2 \\
        \imagepng{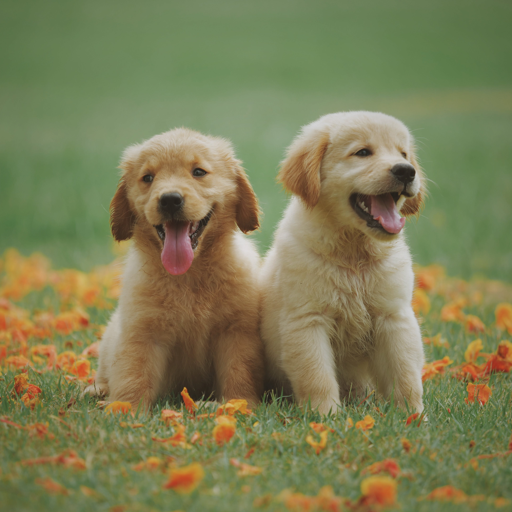} & \imagepng{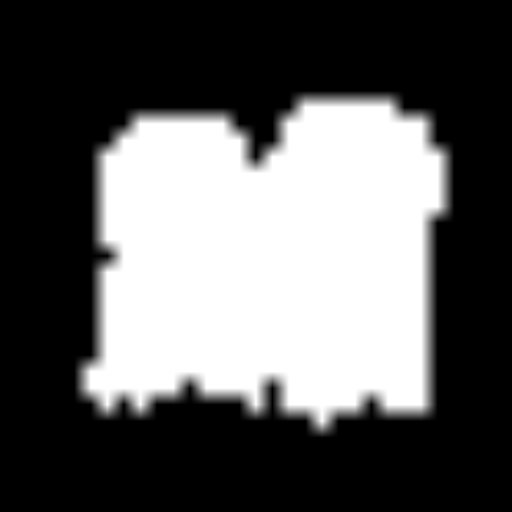} & \imagepng{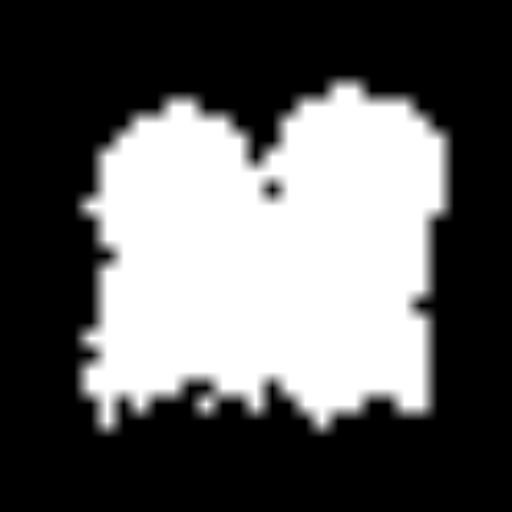}
    \end{tabular}
    \vspace{-2mm}
    \caption{\textbf{Zero-shot foreground-background differentiation using k-means.} Our distilled model produces segmentation masks with less noisy edges.}
\label{fig:downstream}
\end{figure}

\textbf{Other Downstream Tasks} Our distillation and 3D augmentation protocol is tailored for (semantic) correspondence, but we also explore two downstream tasks: foreground segmentation with k-means and ImageNet-1k classification with k-NN probing. As shown in \cref{fig:downstream}, our method qualitatively improves foreground segmentation and slightly outperforms DINOv2 on the DUTS saliency dataset (+0.01\%). However, our model does not perform as well as DINOv2 on the ImageNet k-NN probing (-21.74\%), suggesting that our model focuses more on local rather than global information.

\begin{table}
\centering
\resizebox{0.8\columnwidth}{!}{%
\begin{tabular}{lcc}
\toprule
\textbf{Method} & \textbf{$\text{PCK}_\text{bbox}\mathbf{@0.1}$} & \textbf{Trainable Params} \\
\midrule
Full Fine-tuning & \textbf{62.12} & 86.88M \\
\midrule
Linear Head & 58.96 & \underline{560K} \\
LoRA & \underline{61.77} & \textbf{290K} \\
\bottomrule
\end{tabular}
}
\caption{\textbf{Ablation of two different methods for parameter-efficient fine-tuning, evaluated on SPair-71k.} The best performances are \textbf{bold}, while the second-best are \underline{underlined}.}
\label{tbl:experiments_peft}
\end{table}

\begin{table}
\centering
\resizebox{\columnwidth}{!}{%
\begin{tabular}{llcc}
\toprule
\textbf{Dataset} & \textbf{Method} & \textbf{$\text{PCK}_\text{img}\mathbf{@0.1}$} & \textbf{$\text{PCK}_\text{bbox}\mathbf{@0.1}$} \\
\midrule
SPair-71k & Source sampling & \underline{69.92} & \underline{60.22} \\
& Full sampling & \textbf{71.67} & \textbf{62.37} \\
\midrule
COCO & Mutual-NN sampling & \underline{70.35} & \underline{61.24} \\
& Full sampling & \textbf{71.06} & \textbf{62.02} \\
\bottomrule
\end{tabular}
}
\caption{\textbf{Ablation of three different point sampling methods, trained on two different datasets and evaluated on SPair-71k.} Input resolutions for DINOv2 and SDXL Turbo are $434 \times 434$ and $980 \times 980$ respectively. The best performances are \textbf{bold}, while the second-best are \underline{underlined}.}
\label{tbl:experiments_point_sampling}
\end{table}

\begin{table}
\centering
\resizebox{\columnwidth}{!}{%
\begin{tabular}{llcc}
\toprule
\textbf{Dataset} & \textbf{Method} & \textbf{$\text{PCK}_\text{img}\mathbf{@0.1}$} & \textbf{$\text{PCK}_\text{bbox}\mathbf{@0.1}$} \\
\midrule
SPair-71k & Manually selected pairs & \textbf{71.67} & \textbf{62.37} \\
& Random category pairs & 70.11 & 60.76 \\
& Random pairs & 68.50 & 59.50 \\
& Retrieval pairs & \underline{70.61} & \underline{61.59} \\
\midrule
COCO & Random category pairs & \underline{70.99} & \underline{61.71} \\
& Retrieval pairs & \textbf{71.06} & \textbf{62.02} \\
\bottomrule
\end{tabular}
}
\vspace{-2mm}
\caption{\textbf{Ablation of four different image sampling methods, trained on two different datasets and evaluated on SPair-71k.} Input resolutions for DINOv2 and SDXL Turbo are $434 \times 434$ and $980 \times 980$ respectively. The best performances are \textbf{bold}, while the second-best are \underline{underlined}.}
\vspace{-1em}
\label{tbl:experiments_image_sampling}
\end{table}

\begin{table}
\centering
\resizebox{\columnwidth}{!}{%
\begin{tabular}{llcc}
\toprule
\textbf{Dataset} & \textbf{Method} & \textbf{$\text{PCK}_\text{img}\mathbf{@0.1}$} & \textbf{$\text{PCK}_\text{bbox}\mathbf{@0.1}$} \\
\midrule
SPair-71k & Full sampling & 71.67 & 62.37 \\
&\quad + Timestep ensemble & 71.92 & 62.57 \\
&\quad + Window soft-argmax & 72.00 & 63.85 \\
&\quad + Pose-align & \underline{76.74} & \underline{68.40} \\
&\quad + 3D (ours) & \textbf{77.54} & \textbf{69.09} \\
\midrule
COCO & Retrieval pairs & 71.06 & 62.02 \\
&\quad + Window soft-argmax & \underline{72.36} & \underline{63.35} \\
&\quad + 3D (ours) & \textbf{72.96} & \textbf{63.70} \\
\bottomrule
\end{tabular}
}
\vspace{-2mm}
\caption{\textbf{Ablation of different enhancement techniques applied to our method, trained on two different datasets and evaluated on SPair-71k.} The best performances are \textbf{bold}, while the second-best are \underline{underlined}.}
\label{tbl:experiments_enhance}
\end{table}

\begin{table}
\centering
\resizebox{\columnwidth}{!}{%
\begin{tabular}{lccc}
\toprule
\textbf{Method} & \textbf{\#img/sec} & \textbf{$\text{PCK}_\text{img}\mathbf{@0.1}$} & \textbf{$\text{PCK}_\text{bbox}\mathbf{@0.1}$} \\
\midrule
\cite{zhang2024telling} (w/o pre-train) & \underline{0.4} & - & \textbf{82.9} \\
\midrule
Retrieval pairs & \textbf{28.6} & 71.06 & 62.02 \\
\quad + Supervised fine-tuning &  & 83.39 & 75.24 \\
\quad + Window soft-argmax &  & 84.73 & 77.06 \\
\quad + Dropout &  & \textbf{87.40} & \underline{80.19} \\
\bottomrule
\end{tabular}
}
\vspace{-2mm}
\caption{\textbf{Ablation study on our supervised fine-tuning setting, trained and evaluated on SPair-71k.} The best performances are \textbf{bold}, while the second-best are \underline{underlined}.}
\vspace{-1em}
\label{tbl:experiments_finetune}
\end{table}
\section{Limitations}
Despite these significant contributions, our method still has several limitations that need to be addressed in future work. One major limitation of our unsupervised model is the left-right ambiguity that arises when attempting to differentiate between symmetrical or repeated structures. This issue leads to inaccuracies in differentiating ambiguous parts, e.g. the left and right paws of a cat. Current works that try to address this issue, are all dependent on additional information either in the form of viewpoint information, segmentation masks, or keypoint-specific labels \cite{mariotti2023improving, zhang2024telling}.
Another limitation of all current models is the handling of extreme deformations. For example, the legs of a cat can appear in highly varied positions and angles or even overlap, which can challenge the model's ability to accurately correspond these parts across different images. Current state-of-the-art methods \cite{zhang2023tale, mariotti2023improving, zhang2024telling} still struggle with such deformations, and enhancing the model's robustness to extreme deformations will be essential for improving its applicability to a wider range of real-world images.

\section{Conclusion}
In summary, we successfully distill the knowledge from two large vision foundation models into a smaller student model to improve efficiency without sacrificing performance. Specifically, we leverage the complementary strengths of SDXL Turbo and DINOv2, to distill a more efficient model in terms of parameter count and throughput.
Moreover, we introduce a novel unsupervised fine-tuning method that incorporates 3D data augmentation, enabling the model to achieve new state-of-the-art performance without the need for labeled data.
Despite having fewer parameters and lower input image resolution, our approach demonstrates performance superior to state-of-the-art models across unsupervised, weakly-supervised, and supervised scenarios. This makes our distilled model suitable for resource-constrained environments and real-time applications, while also reducing computational demand during further training.

\section*{Acknowledgements}
\noindent This project has been supported by the German Federal Ministry for Economic Affairs and Climate Action within the project “NXT GEN AI METHODS – Generative Methoden für Perzeption, Prädiktion und Planung”, the bidt project KLIMA-MEMES, and the German Research Foundation (DFG) project 421703927. The authors gratefully acknowledge the Gauss Center for Supercomputing for providing compute through the NIC on JUWELS at JSC and the HPC resources supplied by the Erlangen National High Performance Computing Center (NHR@FAU funded by DFG).

{\small
\bibliographystyle{ieee_fullname}
\bibliography{egbib}
}

\onecolumn
\renewcommand{\thesubsection}{\Alph{subsection}}
\section*{Appendix}
\subsection{Softmax Temperature Ablation}
\begin{figure}[H]
\centering
\includegraphics[width=.5\textwidth]{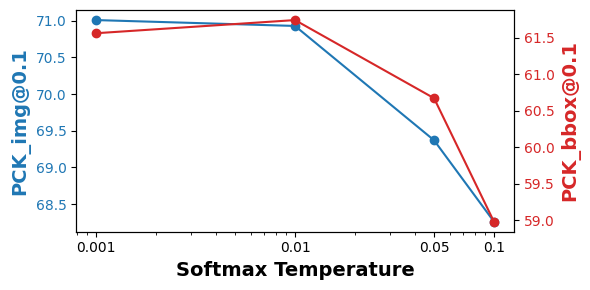}
\caption{\textbf{Ablation of the softmax temperature parameter $\tau$, evaluated on SPair-71k.} Trained for 10 epochs on COCO with retrieval sampling.}
\label{fig:experiments_softmax_temperature}
\end{figure}

In \cref{fig:experiments_softmax_temperature}, we ablate different values for $\tau$ for $\sigma_\tau(\cdot)$. Higher values result in smoother distributions, while lower values result in sharper distributions. We show that the sweet-spot for this parameter is at $0.01$.

\subsection{3D Threshold Ablation}
\begin{figure}[t]
    \centering
    \includegraphics[width=0.8\textwidth]{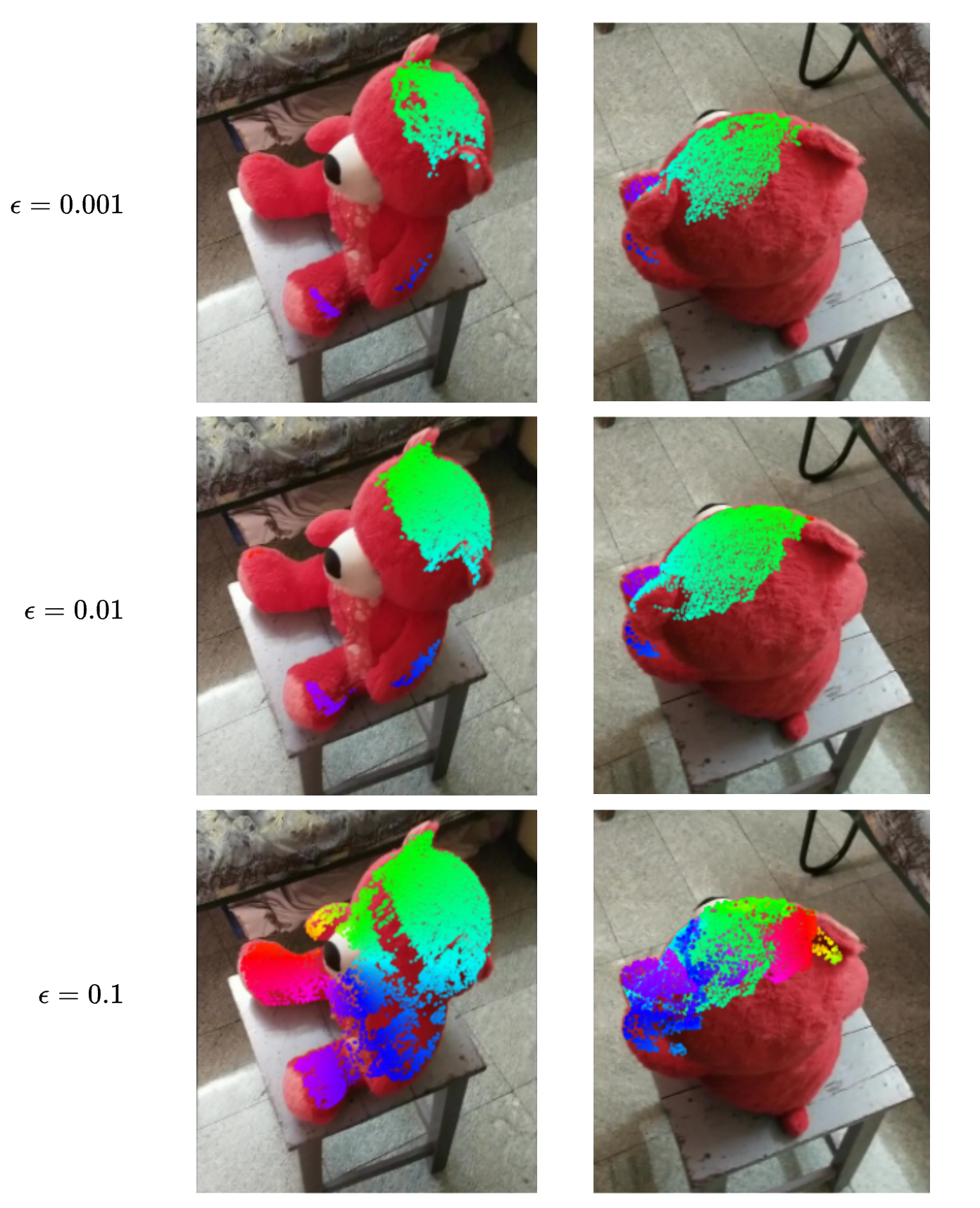}
    \caption{\textbf{The effect of the 3D data threshold parameter $\epsilon$.}}
    \label{fig:3d_threshold}
\end{figure}

\cref{fig:3d_threshold} shows the effect of the threshold parameter $\epsilon$ on our 3D data augmentation method. Smaller values tend to exclude more points that are on the visible surface, whereas larger values tend to include too many points that are not on the visible surface. We set this parameter to $0.01$ as it shows a good balance between those to extremes.

\subsection{Model Analysis}
In \cref{tbl:diff} and \cref{tbl:vit} we ablate different Diffusion- and ViT-based models to find the best combination. We show that SDXL Turbo and DINOv2 (vitb14) with registers are the best performing models in our evaluation. In \cref{tbl:combinations} we ablate different combinations of models to find the best performing setting. We show that the combination of the best performing models in the individual ablation, are also the best performing combination in general. Increasing the input resolution and adding additional layers further boosts the performance. In \Cref{tbl:different_teacher} we ablate the use of an additional strong teacher model, namely CLIP. However, we did not find any improvement in adding this model to the teacher ensemble.\\

\begin{table}[h]
\centering
\renewcommand{\arraystretch}{1.4}
\rowcolors{2}{lightgray!25}{white} 
\begin{tabular}{lcccccc}
\toprule
\textbf{Model} & \textbf{SPair-71K} & \textbf{PF-WILLOW} & \textbf{CUB-200} & \textbf{S} & \textbf{T} & \textbf{L} \\
\toprule
SD1.5 & $66.11 / 56.24$ & $86.58 / 73.60$ & $90.58 / 79.18$ & \(768^2\) & 201 & 5 \\

SD2.1 & $65.29 / 57.87$ & $87.18 / 74.83$ & $88.63 / 78.23$ & \(768^2\) & 261 & 8 \\

SDXL Base & $64.02 / 55.52$ & \underline{$88.37 / 76.49$} & $92.39 / 84.20$ & \(768^2\) & 101 & 1 \\

SDXL Base & $65.64 / 57.87$ & $88.58 / 76.30$ & $92.41 / 84.20$ & \(1024^2\) & 201 & 1 \\

LCM-XL & $62.9 / 54.5$ & $86.52 / 73.81$ & $92.59 / 84.40$ & \(768^2\) & 64 & 1 \\

SDXL Turbo & \underline{$67.26 / 58.54$} & $\mathbf{89.59 / 77.76}$ & \underline{$93.54 / 85.57$} & \(768^2\) & 101 & 1 \\

SDXL Turbo & $\mathbf{67.40 / 59.50}$ & $88.48 / 76.44$ & $\mathbf{93.35 / 85.72}$ & \(1024^2\) & 101 & 1 \\
\bottomrule
\end{tabular}
\caption{\textbf{The performance of different diffusion-based models evaluated on different datasets.} Values are measured in PCK@0.1 (img/bbox), per keypoint and averaged over all keypoints. S: Size of the input image, T: Timestep, L: Layer. Prompt for all models: ``a photo of a [category]".}
\label{tbl:diff}
\end{table}

\begin{table}[t]
\centering
\renewcommand{\arraystretch}{1.4}
\rowcolors{2}{lightgray!25}{white} 
\begin{tabular}{lccccc}
\toprule
\textbf{Model} & \textbf{SPair-71K} & \textbf{PF-WILLOW} & \textbf{CUB-200} & \textbf{R} & \textbf{L} \\
\toprule
CLIP \tiny{(ViT-L-14)} & $47.05 / 37.05$ & $73.51 / 57.67$ & $82.31 / 67.86$ & \(336^2\) & 11 \\

MAE \tiny{(ViT-L-14)} & $33.26 / 23.99$ & $73.04 / 56.54$ & $64.25 / 45.04$ & \(224^2\) & 26 \\

ZoeDepth & $12.80 / 6.63$ & $38.47 / 25.93$ & $22.90 / 9.75$ & \(512 \times 384\) & 10 (BeiT) \\

I-JEPA \tiny{(ViT-H-16 448)} & $51.88 / 44.78$ & $- / -$ & $- / -$ & \(448^2\) & 31 \\

DINOv1 \tiny{(ViT-S-8)} & $46.69 / 35.92$ & $61.66 / 47.99$ & $84.06 / 70.09$ & \(224^2\) & 9 \\

DINOv2 \tiny{(ViT-B-14)} & \underline{$67.45 / 57.69$} & $\mathbf{84.14 / 68.78}$ & \underline{$94.54 / 85.90$} & \(840^2\) & 11 \\

DINOv2R \tiny{(ViT-B-14)} & $\mathbf{69.10 / 58.83}$ & \underline{$83.07 / 67.38$} & $\mathbf{94.61 / 85.90}$ & \(840^2\) & 11 \\
\bottomrule
\end{tabular}
\caption{\textbf{The performance of different ViT-based models evaluated on different datasets.} Values are measured in PCK@0.1 (img/bbox), per keypoint and averaged over all keypoints. S: Size of the input image, L: Layer.}
\label{tbl:vit}
\end{table}

\begin{table}[t]
\centering
\renewcommand{\arraystretch}{1.1}
\rowcolors{2}{lightgray!25}{white} 
\begin{tabular}{lcccccc}
\toprule
\textbf{Model} & \textbf{SPair-71K} & \textbf{PF-WILLOW} & \textbf{CUB-200} & \textbf{S} & \textbf{T} & \textbf{L} \\
\toprule
\begin{tabular}{@{}l}SD1.5 +\\ DINOv2\end{tabular} & $71.57 / 62.03$ & $89.02 / 75.94$ & $94.43 / 85.27$ & \(840^2\) & 201 & \(5\) + \(11\) \\

\begin{tabular}{@{}l}SD1.5 +\\ DINOv2\end{tabular} & $71.38 / 62.08$ & $88.84 / 75.70$ & $94.24 / 85.69$ & \(840^2\) & 201 & \(3, 7, 11\) + \(11\) \\

\begin{tabular}{@{}l}SD1.5 +\\ DINOv2\end{tabular} & \underline{$71.67 / 63.08$} & $88.43 / 74.84$ & $94.55 / 86.25$ & \(960^2\) & 100 & \(3, 7, 11\) + \(11\) \\

\begin{tabular}{@{}l}SDXL Turbo +\\ DINOv2\end{tabular} & $70.90 / 61.88$ & $\mathbf{89.77 / 76.62}$ & \underline{$94.89 / 86.45$} & \(840^2\) & 101 & \(1\) + \(11\) \\

\begin{tabular}{@{}l}SDXL Turbo +\\ DINOv2\end{tabular} & $71.21 / 62.79$ & $88.03 / 74.76$ & $94.22 / 85.81$ & \(840^2\) & 101 & \(1, 4, 7\) + \(11\) \\

\begin{tabular}{@{}l}SDXL Turbo +\\ DINOv2\end{tabular} & $\mathbf{71.77 / 63.29}$ & \underline{$89.36 / 75.98$} & $\mathbf{94.83 / 87.43}$ & \(980^2\) & 101 & \(1\) + \(11\) \\
\bottomrule
\end{tabular}
\caption{\textbf{The performance of different combinations of models and layers evaluated on different datasets.} Values are measured in PCK@0.1 (img/bbox), per keypoint and averaged over all keypoints. With DINOv2, we mean DINOv2 (ViT-B-14) with registers. S: Size of the input image, T: Timestep, L: Layer. Prompt for all models: ``a photo of a [category]".}
\label{tbl:combinations}
\end{table}

\begin{table}[H]
\centering
\begin{tabular}{lcc}
\toprule
\textbf{Method} & \textbf{$\text{PCK}_\text{img}\mathbf{@0.1}$} & \textbf{$\text{PCK}_\text{bbox}\mathbf{@0.1}$} \\
\midrule
DINOv2 + SD & \textbf{71.77} & \textbf{63.29} \\
DINOv2 + SD + CLIP & \underline{68.87} & \underline{60.17} \\
\bottomrule
\end{tabular}
\caption{\textbf{Performance on SPair-71k for different teachers.} Adding CLIP to the teacher ensemble does not improve performance.}
\label{tbl:different_teacher}
\end{table}

\subsection{Foreground Segmentation}
\begin{figure}[t]
    \centering
    \includegraphics[width=\textwidth]{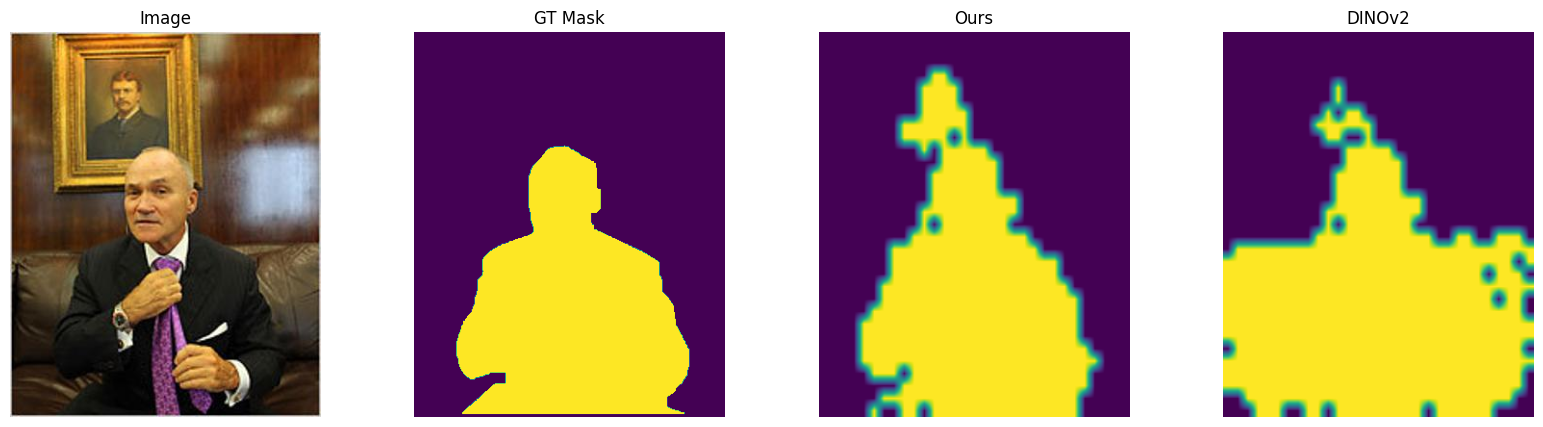}
    \includegraphics[width=\textwidth]{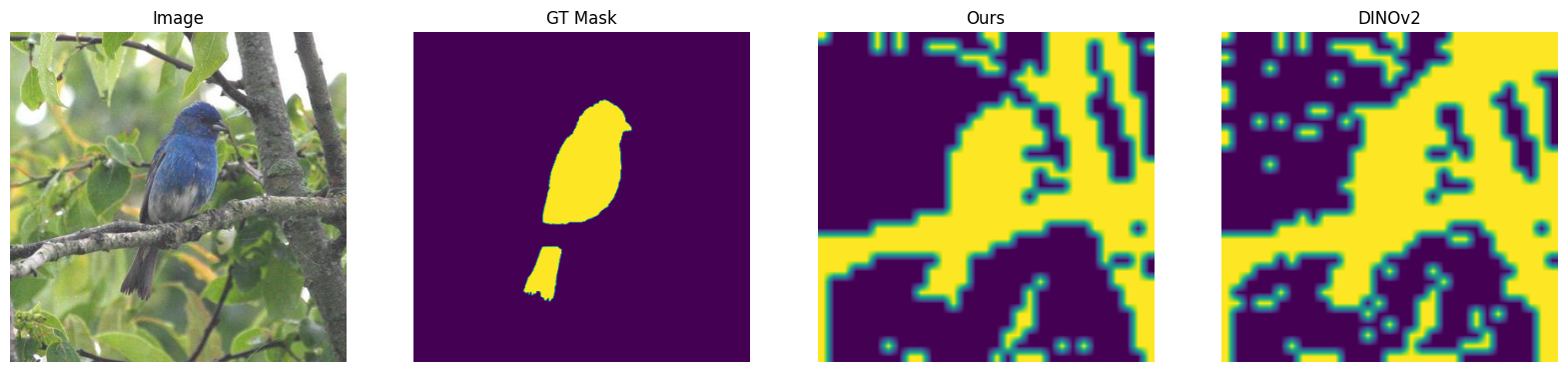}
    \includegraphics[width=\textwidth]{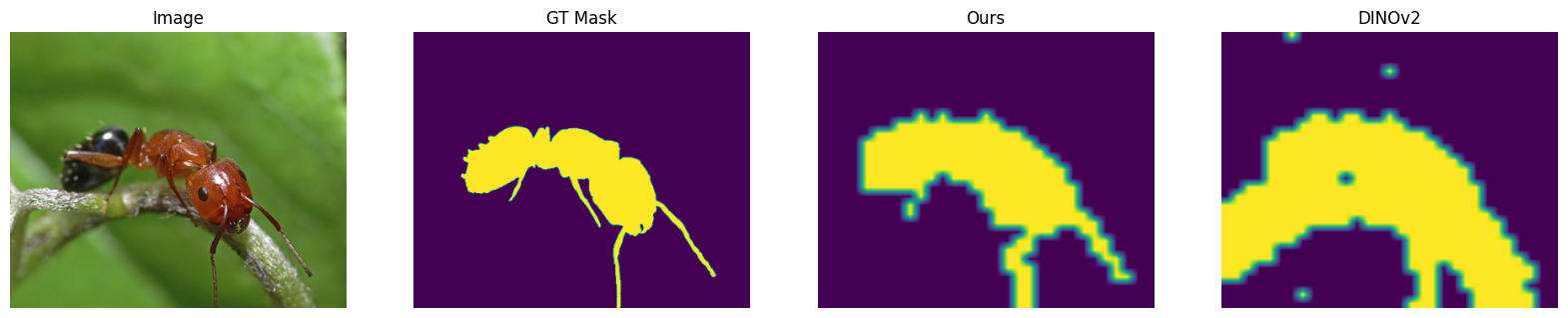}
    \includegraphics[width=\textwidth]{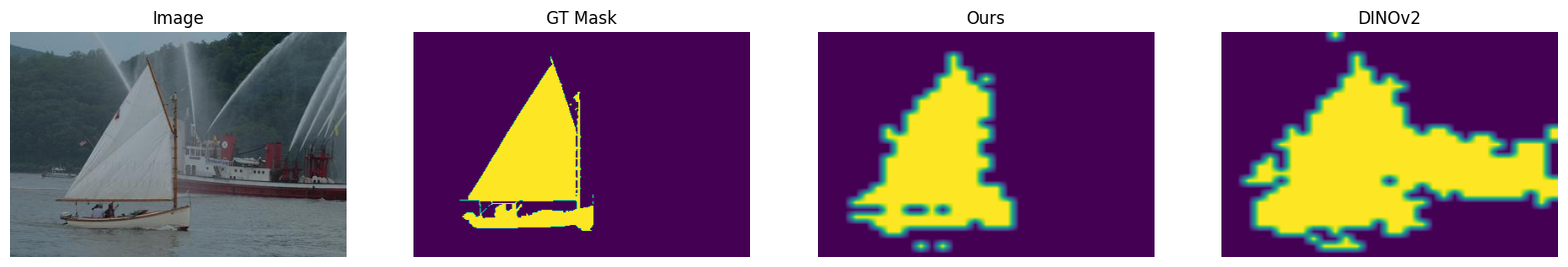}
    \caption{\textbf{Examples of the improved foreground/background segmentation masks with our model.}}
    \label{fig:fgbg}
\end{figure}

We assess our model on other downstream tasks, including zero-shot foreground/ background segmentation. The examples in \Cref{fig:fgbg} show a marginal improvement in mask quality. The masks generated with our model are slightly less noisy compared to the baseline DINOv2 model.

\end{document}